\documentclass[lettersize,journal]{IEEEtran}
\usepackage{amsmath,amssymb,amsthm,graphicx, comment}
\usepackage{xcolor}
\newtheorem{proposition}{Proposition}
\newtheorem{theorem}{Theorem}
\newtheorem{remark}{Remark}
\newtheorem{definition}{Definition}

\newcommand{\norm}[1]{\left\lVert#1\right\rVert}

\usepackage{amsfonts}
\usepackage{algorithmic}
\usepackage{algorithm}
\usepackage{array}
\usepackage[caption=false,font=normalsize,labelfont=sf,textfont=sf]{subfig}
\usepackage{textcomp}
\usepackage{stfloats}
\usepackage{url}
\usepackage{verbatim}
\usepackage{graphicx}
\usepackage{cite}
\begin{document}

\title{Directional Bounded-Confidence Matrix Alignment in Vicsek Flocking}

\author{Hossein B. Jond
\thanks{This work was funded by the Czech Science Foundation (GA\v{C}R) under research project no.\ 23-07517S and the European Union under the project  Robotics and Advanced Industrial Production  (reg.\ no.\ CZ.02.01.01/00/22\_008/0004590). }
\thanks{H. B. Jond is with the Department of Cybernetics, Faculty of Electrical Engineering, Czech Technical University in Prague, Czech Republic (e-mail: hossein.barghi.jond@cvut.cz).}
}

\maketitle

\begin{abstract}
This paper proposes a directional bounded-confidence variant of the discrete-time Vicsek multiagent model, in which pairwise heading alignment is modulated by a state-dependent, positive semidefinite matrix that independently gates orthogonal radial and tangential components in the instantaneous line-of-sight frame. We analyze the normalized, matrix-weighted system dynamics, proving a uniform non-vanishing bound on proposal vectors that guarantees global well-posedness across arbitrary network states. For local small-spread states under complete metric interaction, we prove linear geometric contraction of the heading diameter and establish an offline-verifiable spatial bootstrap condition that guarantees metric contact preservation and coincidence avoidance a priori. A counterexample demonstrates invariant polarized states on connected matrix graphs, delineating systemic structural limits. Numerical experiments against unfiltered Vicsek averaging, scalar tied confidence, trace-matched scalar baselines, and restricted-angle rules demonstrate component-selective filtering without rejected-channel leakage, along with improved fixed-bound alignment on multi-seed benchmarks across varying population densities.
\end{abstract}


\section{Introduction}
\IEEEPARstart{M}{ultiagent} coordination and distributed consensus over complex network topologies serve as foundational pillars for modern cyber-physical systems, autonomous robotic swarms, sensor networks, and decentralized control architectures~\cite{4118472}. In classical multiagent control frameworks, interaction dynamics rely primarily on scalar coupling weights, implicitly assuming that all spatial, directional, and operational state components share identical interaction gains across channels~\cite{jadbabaie2003}. However, real-world networked systems—such as autonomous aerial vehicle (UAV) formations operating under directional line-of-sight (LOS) constraints, sensor swarms with asymmetric field-of-view limits, and decentralized belief-action systems—frequently exhibit component-dependent interdependencies and spatial orientation constraints~\cite{Shang2023_JSYST}. Standard isotropic scalar rules cannot capture these anisotropic coupling dynamics, as they enforce uniform gain scaling across all state dimensions regardless of directional orientation.

To address these limitations, matrix-weighted consensus theory replaces scalar edge weights with state-dependent matrix operators~\cite{trinh2018,Tran2021,Shang2023_JSYST}. Matrix interaction maps enable independent filtering, directional gating, and cross-channel coupling across orthogonal state dimensions, allowing multiagent systems to execute complex anisotropic control protocols~\cite{trinh2018,Tran2021}. Recent advances in matrix-scaled consensus further characterize the mathematical foundations of multi-dimensional state transformation spaces, anisotropic communication graphs, and matrix-weighted Graph Laplacians~\cite{Trinh2024_TNSE}. 

Parallel to control-theoretic developments, self-organization and flocking kinetics at the interface of active matter physics and robotics offer bio-inspired paradigms for resilient swarm control~\cite{vasarhelyi2018flock,boldrer2026distributed,jond2026}. Among foundational swarm kinetics, the discrete-time model introduced by Vicsek \textit{et al.}~\cite{vicsek1995} remains a bedrock paradigm for analyzing emergent consensus, phase transitions, and collective alignment in active particle ensembles. To capture realistic cognitive and perceptual constraints, bounded-confidence mechanisms refine classic flocking dynamics by enforcing state-dependent compatibility criteria: agents interact only when their state separation falls within a specified confidence threshold~\cite{hegselmann2002,deffuant2000mixing}.

Existing restricted-angle bounded-confidence Vicsek variants~\cite{romensky2014,romenskyy2018} rely exclusively on scalar compatibility tests that accept or reject a neighbor's complete heading vector based on total scalar angular separation. This modification fundamentally alters global system dynamics, frequently triggering multi-group fragmentation and stable polarized clustering rather than unified flocking~\cite{song2026}. Other variants model non-reciprocal visual perception cones~\cite{gaur2026} or introduce scalar weights to accelerate consensus convergence~\cite{ZOU20133329}. Nevertheless, all such scalar bounded-confidence formulations remain intrinsically isotropic: they cannot independently gate orthogonal spatial channels—such as separating relative approach motion along the instantaneous LOS from transverse angular motion.

In this paper, we bridge matrix-weighted network consensus with bounded-confidence Vicsek kinetics by introducing a directional, matrix-weighted bounded-confidence multiagent model. By decomposing pairwise relative headings into orthogonal radial and tangential components within the instantaneous LOS frame, we construct a state-dependent, positive semidefinite (PSD) matrix operator that independently gates orthogonal motion channels. Under this systems framework, two agent pairs with identical total angular separation receive distinct matrix corrections depending on their relative spatial bearings. A scalar rule cannot capture this directional selectivity, and a bearing-dependent scalar gain cannot independently gate orthogonal channels. Benchmarking against a trace-matched scalar baseline derived from the same component gates is therefore essential to isolate channel-selective matrix action from scalar gain variations.

Designing such state-dependent matrix interaction maps introduces distinct systems-level analytical challenges. Because matrix projections onto instantaneous LOS frames are non-commutative and state-dependent, the resulting interaction Laplacian breaks standard pairwise momentum conservation. Furthermore, coupling state-dependent matrix gating with discrete-time vector normalization creates non-Lipschitz update manifolds that risk division-by-zero singularities if proposal vector norms vanish. Addressing these systems-level hurdles requires a rigorous verification of well-posedness, spatial drift, and structural convergence limits.

The main contributions of this paper are fourfold. First, we establish a global well-posedness guarantee proving that proposal vectors never vanish, ensuring non-singular state normalization across arbitrary network states. Second, a local flocking theorem proves linear geometric contraction of the heading diameter under complete interaction. Third, an a priori spatial bootstrap condition provides an offline-verifiable initial constraint guaranteeing edge preservation and collision avoidance. Fourth, an exact invariant counterexample reveals structural limits by proving that matrix graph connectivity alone is insufficient for consensus. Finally, multi-seed comparative benchmarking across swarms ranging from 10 to 200 agents confirms zero rejected-channel leakage and superior fixed-bound alignment over standard scalar and restricted-angle baselines.

The remainder of this paper is organized as follows. Section~\ref{sec:model} formalizes the directional bounded-confidence model, state space, and normalization updates. Section~\ref{sec:analysis} presents the theoretical analysis, proving well-posedness, local flocking contraction, a priori bootstrap conditions, and structural limits on matrix graphs. Section~\ref{sec:numerical} validates the analytical bounds through comprehensive numerical simulations, multi-seed population sweeps, and baseline comparisons. Section~\ref{sec:conclusion} concludes the paper and discusses future directions.

\section{Directional Vicsek Model} \label{sec:model}
\subsection{State Space and Anisotropic Geometry}
Consider a multi-agent network of $n \ge 2$ agents evolving in discrete time $k \in \mathbb{N}_0$. The state of agent $i \in \{1, \dots, n\}$ at time step $k$ is defined by its position $\mathbf p_i^k \in \mathbb R^2$ and unit heading vector $\mathbf u_i^k \in \mathbb S^1 \subset \mathbb R^2$. Agents move at a constant nominal speed $v_0 > 0$, yielding physical velocity $\mathbf v_i^k = v_0 \mathbf u_i^k$.

For any distinct pair of agents $i \neq j$, define the relative position vector, unit LOS vector, and relative heading difference at step $k$ as:
\begin{equation*}
 \mathbf r_{ij}^k = \mathbf p_j^k - \mathbf p_i^k, \quad \mathbf e_{ij}^k = \frac{\mathbf r_{ij}^k}{\norm{\mathbf r_{ij}^k}}, \quad \mathbf w_{ij}^k = \mathbf u_j^k - \mathbf u_i^k. 
\end{equation*}
In the instantaneous pairwise LOS frame, define the symmetric orthogonal projection operators onto the radial and tangential subspace channels, respectively:
\begin{equation*}
 \mathbf P^r_{ij} = \mathbf e_{ij} (\mathbf e_{ij})^{\mathsf T}, \qquad \mathbf P^t_{ij} = \mathbf I_2 - \mathbf P^r_{ij}, 
\end{equation*}
where $\mathbf I_2 \in \mathbb R^{2 \times 2}$ is the identity matrix and time index $k$ is suppressed for brevity. The radial vector component $\mathbf P^r_{ij} \mathbf w_{ij}$ measures relative velocity along the LOS divided by speed $v_0$, while the tangential component $\mathbf P^t_{ij} \mathbf w_{ij}$ measures transverse relative velocity divided by $v_0$.

\begin{figure}[t]
\centering\includegraphics[width=0.8\columnwidth]{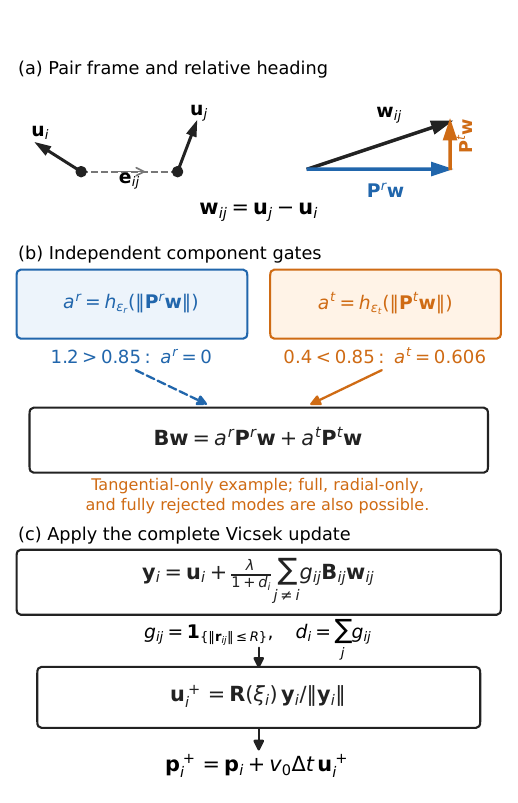}
\caption{Schematic of the state-dependent directional-confidence alignment mechanism. (a)~Relative heading vector $\mathbf w_{ij} = \mathbf u_j - \mathbf u_i$ decomposed into orthogonal radial ($\mathbf P^r_{ij}\mathbf w_{ij}$) and tangential ($\mathbf P^t_{ij}\mathbf w_{ij}$) components in the instantaneous LOS frame $\mathbf e_{ij} = \mathbf r_{ij}/\norm{\mathbf r_{ij}}$. (b)~Independent gating functions $h_{\epsilon_r}$ and $h_{\epsilon_t}$ yield anisotropic edge matrix $\mathbf B_{ij}$, illustrated here for a tangential-only filtering scenario ($\epsilon_r = \epsilon_t = 0.85$) where the radial correction is completely suppressed. (c)~Filtered relative corrections enter the metric-masked residual sum, followed by state normalization $\mathcal N_i(\mathbf y_i^k)$, uniform angular noise rotation $\mathbf R(\xi_i^k)$, and spatial streaming. All geometric projections and confidence matrices are evaluated synchronously at step $k$; superscript $+$ denotes the updated state at $k+1$.} 
\label{fig:mechanism}
\end{figure}

\subsection{State-Dependent Anisotropic Matrix Confidence}
To filter relative heading interactions independently along radial and tangential channels, let $h_\epsilon : [0, \infty) \to [0, 1]$ be a compactly supported $C^1$ scalar gating function parameterized by threshold $\epsilon > 0$:
\begin{equation*}
 h_\epsilon(s) = \left[ \max\left\{0, 1 - \frac{s^2}{\epsilon^2}\right\} \right]^2.
\end{equation*}
Using dimensionless channel heading thresholds $\epsilon_r, \epsilon_t > 0$, we define the state-dependent radial and tangential edge gains:
\begin{align}
 a^r_{ij} &= h_{\epsilon_r}\left( \norm{\mathbf P^r_{ij} \mathbf w_{ij}} \right), \quad a^t_{ij} = h_{\epsilon_t}\left( \norm{\mathbf P^t_{ij} \mathbf w_{ij}} \right), \notag \\
 \mathbf B_{ij} &= a^r_{ij} \mathbf P^r_{ij} + a^t_{ij} \mathbf P^t_{ij}. \label{eq:block}
\end{align}
By construction, the interaction block matrix is symmetric ($\mathbf B_{ij} = \mathbf B_{ji}$) and PSD with bounded spectrum $0 \preceq \mathbf B_{ij} \preceq \mathbf I_2$. As illustrated in Fig.~\ref{fig:mechanism}, the interaction operates in one of four active channel modes depending on whether $a^r_{ij}$ and $a^t_{ij}$ exceed zero: \emph{full}, \emph{radial-only}, \emph{tangential-only}, or \emph{rejected}.

Note that since $\norm{\mathbf w_{ij}} = 2\left|\sin\left(\frac{\theta_j - \theta_i}{2}\right)\right|$ for heading angles $\theta_i, \theta_j$, a chord threshold $\epsilon \le 2$ maps to a maximum angular support bound of $2\arcsin(\epsilon/2)$ rad. If physical velocity thresholds $\epsilon_{v,r}, \epsilon_{v,t}$ are specified instead, dimensionless thresholds are obtained directly via $\epsilon_r = \epsilon_{v,r} / v_0$ and $\epsilon_t = \epsilon_{v,t} / v_0$.

\subsection{Residual Alignment and Discrete-Time Dynamics}
Spatial connectivity is governed by an unweighted, metric range mask with interaction radius $R > 0$. Let $g_{ij}^k = \mathbf 1_{\{\norm{\mathbf r_{ij}^k} \le R\}}$ for $i \neq j$, with $g_{ii}^k = 0$, and define the metric degree $d_i^k = \sum_{j=1}^n g_{ij}^k$.

To guarantee that perfectly aligned agents undergo zero heading correction regardless of spatial orientation, we formulate the alignment map in residual difference form. The proposed update rule proceeds synchronously for each agent $i$:
\begin{align}
 \mathbf y_i^k &= \mathbf u_i^k + \frac{\lambda}{1 + d_i^k} \sum_{j \neq i} g_{ij}^k \mathbf B_{ij}^k (\mathbf u_j^k - \mathbf u_i^k), \label{eq:proposal} \\
 \mathbf u_i^{k+1} &= \mathbf R(\xi_i^k) \mathcal N_i(\mathbf y_i^k), \label{eq:heading} \\
 \mathbf p_i^{k+1} &= \mathbf p_i^k + v_0 \Delta t \, \mathbf u_i^{k+1}, \label{eq:position}
\end{align}
where $0 < \lambda \le 1$ is a step-size (relaxation) constant, $\Delta t > 0$ is the sampling interval, $\mathbf R(\xi) \in \mathrm{SO}(2)$ is a 2D rotation matrix, and $\xi_i^k \sim \mathrm{Unif}[-\eta/2, \eta/2]$ represents independent bounded angular noise. 

The vector normalization operator $\mathcal N_i : \mathbb R^2 \to \mathbb S^1$ is defined as:
\begin{equation*}
 \mathcal N_i(\mathbf y) = \begin{cases} \frac{\mathbf y}{\norm{\mathbf y}}, & \text{if } \mathbf y \neq \mathbf 0, \\ \mathbf u_i^k, & \text{if } \mathbf y = \mathbf 0. \end{cases} 
\end{equation*}
All geometric states ($\mathbf r_{ij}^k, \mathbf w_{ij}^k$) and matrix blocks $\mathbf B_{ij}^k$ are evaluated at time step $k$, and spatial streaming in~\eqref{eq:position} uses the updated heading $\mathbf u_i^{k+1}$.

\begin{remark}
If sampled trajectories encounter an exact pairwise coincidence ($\mathbf p_i^k = \mathbf p_j^k$), the LOS vector $\mathbf e_{ij}^k$ is undefined. At exact coincidence, block~\eqref{eq:block} is replaced by $h_{\epsilon_{\min}}(\norm{\mathbf w_{ij}^k})\mathbf I_2$ with $\epsilon_{\min} = \min(\epsilon_r, \epsilon_t)$, maintaining rotation equivariance. When all gates are bypassed ($\mathbf B_{ij} = \mathbf I_2$) and $\lambda = 1$, formulation~\eqref{eq:proposal}--\eqref{eq:heading} reduces to standard self-inclusive Vicsek neighborhood averaging:
\begin{equation*}
 \mathbf y_i^k = \frac{\mathbf u_i^k + \sum_{j} g_{ij}^k \mathbf u_j^k}{1 + d_i^k}.
\end{equation*}
\end{remark}

\section{Analysis} \label{sec:analysis}
In this section, we analyze the deterministic heading dynamics ($\eta = 0$). We first establish global well-posedness by proving that proposal vectors never collapse to zero for relaxation constants $\lambda < 1/2$. We then provide a complete, step-by-step convergence proof for local heading contraction under complete interaction, followed by an a priori spatial bootstrap condition and a structural counterexample on connected matrix graphs.

\begin{proposition}[Global Well-Posedness]\label{prop:safe}
Under updates~\eqref{eq:proposal}--\eqref{eq:heading} with $\eta = 0$, unit speed $\norm{\mathbf u_i^k} = 1$ is invariant for all $k \ge 0$. Furthermore, $0 < \lambda < 1/2$ is a sufficient condition for global well-posedness on arbitrary heading states and spatial graphs, guaranteeing non-vanishing proposal vectors:
\begin{equation*}
 \norm{\mathbf y_i^k} \ge 1 - \frac{2\lambda d_i^k}{1 + d_i^k} > 1 - 2\lambda > 0, \quad \forall i \in \{1, \dots, n\}.
\end{equation*}
Finally, every synchronized agreement state $\mathbf u_1^k = \dots = \mathbf u_n^k = \mathbf u_\star \in \mathbb S^1$ is an invariant manifold of the heading subsystem.
\end{proposition}
\begin{proof}
Since $0 \preceq \mathbf B_{ij}^k \preceq \mathbf I_2$, the operator norm satisfies $\norm{\mathbf B_{ij}^k} \le 1$. For unit headings $\mathbf u_i, \mathbf u_j \in \mathbb S^1$, the difference norm is bounded by $\norm{\mathbf w_{ij}^k} \le 2$. Applying the norm inequality to the correction term in~\eqref{eq:proposal} yields:
\begin{align*}
\norm{\frac{\lambda}{1+d_i^k}\sum_{j\ne i}g_{ij}^k\mathbf B_{ij}^k\mathbf w_{ij}^k} &\le \frac{\lambda}{1+d_i^k}\sum_{j\ne i}g_{ij}^k\norm{\mathbf B_{ij}^k}\norm{\mathbf w_{ij}^k}\\& \le \frac{2\lambda d_i^k}{1+d_i^k}.
\end{align*}
Applying the reverse triangle inequality to $\mathbf y_i^k = \mathbf u_i^k + \text{correction}$ gives:
\begin{equation*}
\norm{\mathbf y_i^k} \ge \norm{\mathbf u_i^k} - \norm{\text{correction}} \ge 1 - \frac{2\lambda d_i^k}{1+d_i^k}.
\end{equation*}
Since $d_i^k / (1 + d_i^k) < 1$, taking $\lambda < 1/2$ guarantees $\norm{\mathbf y_i^k} > 1 - 2\lambda > 0$. Thus $\mathbf y_i^k \neq \mathbf 0$ everywhere, ensuring that normalization $\mathcal N_i(\mathbf y_i^k) = \mathbf y_i^k / \norm{\mathbf y_i^k}$ is everywhere smooth and well-defined without division-by-zero singularities. When $\mathbf u_i^k = \mathbf u_\star$ for all $i$, every $\mathbf w_{ij}^k = \mathbf 0$, giving $\mathbf y_i^k = \mathbf u_\star$ and confirming manifold invariance.
\end{proof}

As positions approach coincidence ($\mathbf r_{ij} \to \mathbf 0$), the LOS unit vector $\mathbf e_{ij}$ depends on the approach direction $\mathbf n = \lim \mathbf r_{ij}/\norm{\mathbf r_{ij}} \in \mathbb S^1$. Because $\mathbf B_{ij}$ gates projections onto this directional frame, the limit $\lim_{\mathbf p_j \to \mathbf p_i} \mathbf y_i^k$ varies with $\mathbf n$, creating a directional jump discontinuity at exact coincidence $\mathbf p_i = \mathbf p_j$. Nevertheless, as established in Proposition~\ref{prop:safe}, $\lambda < 1/2$ guarantees non-vanishing proposal norms ($\norm{\mathbf y_i^k} > 1 - 2\lambda > 0$) everywhere on $\mathbb R^2 \times \mathbb S^1$, so normalization $\mathcal N_i(\mathbf y_i^k)$ encounters no division-by-zero singularities regardless of agent proximity.

\begin{proposition}\label{prop:small_spread_lambda}
Suppose the heading diameter at step $k$ satisfies $\Delta_k \le \delta$ for a small-spread bound $\delta > 0$. Under updates~\eqref{eq:proposal}--\eqref{eq:position} with gating threshold $\epsilon_{\min} = \min(\epsilon_r, \epsilon_t) > 0$, proposal vectors satisfy $\norm{\mathbf y_i^k} \ge b(\lambda, \delta) > 0$ for all agents $i \in \{1, \dots, n\}$ whenever the step-size $\lambda$ satisfies:
\begin{equation*}
 0 < \lambda < \lambda_{\max}(\delta) \triangleq \frac{1}{\frac{1}{2}\delta^2 + \frac{2\delta^3}{\epsilon_{\min}^2}}. \label{eq:lambda_max_def}
\end{equation*}
In particular, $\lambda_{\max}(\delta)>1$ holds when
$\frac12\delta^2+2\delta^3/\epsilon_{\min}^2<1$.
Under this condition every $0<\lambda\le1$ preserves well-posedness ($\norm{\mathbf y_i^k} \ge b > 0$).
\end{proposition}
\begin{proof}
Decompose $\mathbf y_i^k = \mathbf u_i^k + \frac{\lambda}{1+d_i^k}\sum_{j \neq i} g_{ij}^k(\mathbf u_j^k - \mathbf u_i^k) + \mathbf E_i^k$ as in Theorem~\ref{thm:local}. Taking the inner product with unit vector $\mathbf u_i^k$ and using $(\mathbf u_i^k)^{\mathsf T}\mathbf u_j^k = 1 - \frac{1}{2}\norm{\mathbf w_{ij}^k}^2 \ge 1 - \frac{1}{2}\Delta_k^2$ yields:
\begin{align*}
 (\mathbf u_i^k)^{\mathsf T}\mathbf y_i^k &\ge 1 - \frac{\lambda d_i^k}{1+d_i^k}\left(\frac{1}{2}\Delta_k^2\right) - \norm{\mathbf E_i^k} \nonumber \\
 &\ge 1 - \frac{\lambda d_i^k}{1+d_i^k}\left( \frac{1}{2}\Delta_k^2 + \frac{2\Delta_k^3}{\epsilon_{\min}^2} \right).
\end{align*}
Since $d_i^k / (1+d_i^k) < 1$ and $\Delta_k \le \delta$, we obtain $\norm{\mathbf y_i^k} \ge (\mathbf u_i^k)^{\mathsf T}\mathbf y_i^k \ge 1 - \lambda \left( \frac{1}{2}\delta^2 + \frac{2\delta^3}{\epsilon_{\min}^2} \right) \triangleq b(\lambda, \delta)$. Ensuring $b(\lambda, \delta) > 0$ directly gives condition~\eqref{eq:lambda_max_def}.
\end{proof}

\subsection{Normalization, Conservation, and Non-Conserved Means}
Let $\mathbf L_B^k \in \mathbb R^{2n \times 2n}$ be the symmetric block Laplacian matrix with off-diagonal $2 \times 2$ blocks $-g_{ij}^k \mathbf B_{ij}^k$ and diagonal blocks $\sum_{j \neq i} g_{ij}^k \mathbf B_{ij}^k$. Let $\mathbf D^k = \operatorname{diag}(1+d_i^k) \in \mathbb R^{n \times n}$ and stacked heading vector $\mathbf u^k = \operatorname{col}(\mathbf u_1^k, \dots, \mathbf u_n^k) \in \mathbb R^{2n}$. The stacked proposal vector is:
\begin{equation*}
 \mathbf y^k = \left(\mathbf I_{2n} - \lambda \left( (\mathbf D^k)^{-1} \otimes \mathbf I_2 \right) \mathbf L_B^k\right) \mathbf u^k.
\end{equation*}
Unlike standard linear Laplacian consensus, normalization $\mathcal N_i(\mathbf y_i^k)$ destroys heading mean conservation even when degrees are uniform.

\begin{remark}
Degree scaling $1/(1+d_i^k)$ and vector normalization $\mathcal N_i(\mathbf y_i^k)$ break mean-heading conservation, even on regular graphs. This mirrors normalized active-matter models (e.g., Motsch--Tadmor kinetics~\cite{motsch2011}), where radial projection biases updates toward dense, aligned clusters rather than conserving total system momentum.
\end{remark}

\subsection{Local Flocking Under Complete Interaction}
Define the heading diameter $\Delta_k$ and maximum spatial displacement diameter $X_k$ at step $k$:
\begin{equation*}
 \Delta_k = \max_{i,j} \norm{\mathbf u_i^k - \mathbf u_j^k}, \qquad X_k = \max_{i,j} \norm{\mathbf p_i^k - \mathbf p_j^k}.
\end{equation*}

\begin{theorem}\label{thm:local}
Consider updates~\eqref{eq:proposal}--\eqref{eq:position} with $0 < \lambda \le 1$ and gating threshold $\epsilon_{\min} = \min(\epsilon_r, \epsilon_t) > 0$. Choose a small-spread bound $\delta > 0$ such that the constants:
\begin{align}
 b &= 1 - \frac{\lambda \delta^2}{2} - \frac{2\lambda \delta^3}{\epsilon_{\min}^2} > 0, \notag \\
 q &= \frac{1 - \lambda + \frac{4\lambda \delta^2}{\epsilon_{\min}^2}}{b} < 1, \label{eq:q_def}
\end{align}
are strictly satisfied. If the initial heading spread satisfies $\Delta_0 \le \delta$, pairwise agent positions remain distinct, and the spatial graph is complete ($g_{ij}^k = 1, \, \forall i \neq j$) for all $k \ge 0$, then:
\begin{enumerate}
    \item The heading diameter contracts linearly:
    \begin{equation*}
     \Delta_k \le q^k \Delta_0, \quad \forall k \ge 0.
    \end{equation*}
    \item The spatial diameter obeys:
    \begin{equation*}
     X_k \le X_0 + v_0 \Delta t \, \Delta_0 \frac{q}{1 - q}, \quad \forall k \ge 0.
    \end{equation*}
\end{enumerate}
Consequently, all headings converge to a common limit vector $\mathbf u_\star \in \mathbb S^1$, and all relative pairwise positions converge.
\end{theorem}

\begin{proof}
The proof proceeds in four explicit steps:

\emph{Step 1: Matrix Block Perturbation Bound.}
For any $s \ge 0$, the gating function $h_\epsilon(s) = [\max\{0, 1 - s^2/\epsilon^2\}]^2$ satisfies $0 \le 1 - h_\epsilon(s) \le 2s^2/\epsilon^2$. Under complete interaction with maximum pairwise spread $\Delta_k$, the radial and tangential distances satisfy $\norm{\mathbf P_{ij}^r \mathbf w_{ij}^k} \le \Delta_k$ and $\norm{\mathbf P_{ij}^t \mathbf w_{ij}^k} \le \Delta_k$. Therefore, the matrix deviation from identity is bounded by:
\begin{equation}\label{eq:proof_step1}
 \norm{\mathbf B_{ij}^k-\mathbf I_2}
 =\max_{c\in\{r,t\}}\left|1-h_{\epsilon_c}(\norm{\mathbf P_{ij}^c\mathbf w_{ij}^k})\right|\le 2\Delta_k^2/\epsilon_{\min}^2.
\end{equation}

\emph{Step 2: Decomposition and Perturbation Bounds on Proposal Vectors.}
Under complete interaction, $g_{ij}^k = 1$ for $j \neq i$, so $d_i^k = n - 1$ and $1 + d_i^k = n$. Define the arithmetic mean heading $\overline{\mathbf u}^k = \frac{1}{n} \sum_{j=1}^n \mathbf u_j^k$. We decompose proposal vector $\mathbf y_i^k$ into ideal linear consensus plus a matrix perturbation $\mathbf E_i^k$:
\begin{align}
 \mathbf y_i^k &= \mathbf u_i^k + \frac{\lambda}{n}\sum_{j \neq i}(\mathbf u_j^k - \mathbf u_i^k) + \frac{\lambda}{n}\sum_{j \neq i}(\mathbf B_{ij}^k - \mathbf I_2)\mathbf w_{ij}^k \nonumber \\
 &= (1 - \lambda)\mathbf u_i^k + \lambda \overline{\mathbf u}^k + \mathbf E_i^k, \label{eq:proof_step2_decomp}
\end{align}
where $\mathbf E_i^k = \frac{\lambda}{n}\sum_{j \neq i}(\mathbf B_{ij}^k - \mathbf I_2)\mathbf w_{ij}^k$. Applying bound~\eqref{eq:proof_step1} and $\norm{\mathbf w_{ij}^k} \le \Delta_k$, the perturbation norm is bounded uniformly across all agents by:
\begin{equation}\label{eq:proof_step2_Ebound}
 \norm{\mathbf E_i^k} \le \frac{\lambda(n-1)}{n} \left(\max_{j} \norm{\mathbf B_{ij}^k - \mathbf I_2}\right) \Delta_k \le \frac{2\lambda \Delta_k^3}{\epsilon_{\min}^2}.
\end{equation}

Next, evaluate the projection of proposal $\mathbf y_i^k$ onto unit vector $\mathbf u_i^k$. Since $(\mathbf u_i^k)^{\mathsf T} \mathbf u_j^k = 1 - \frac{1}{2}\norm{\mathbf w_{ij}^k}^2 \ge 1 - \frac{1}{2}\Delta_k^2$, taking inner products yields:
\begin{align}
 (\mathbf u_i^k)^{\mathsf T} \mathbf y_i^k &= (1 - \lambda) + \frac{\lambda}{n}\sum_{j=1}^n (\mathbf u_i^k)^{\mathsf T} \mathbf u_j^k + (\mathbf u_i^k)^{\mathsf T} \mathbf E_i^k \nonumber \\
 &\ge (1 - \lambda) + \lambda\left(1 - \frac{1}{2}\Delta_k^2\right) - \norm{\mathbf E_i^k} \nonumber \\
 &\ge 1 - \frac{\lambda \Delta_k^2}{2} - \frac{2\lambda \Delta_k^3}{\epsilon_{\min}^2}. \label{eq:proof_step2_lowerbound}
\end{align}
Since $\Delta_k \le \delta$, bound~\eqref{eq:proof_step2_lowerbound} establishes that $\norm{\mathbf y_i^k} \ge (\mathbf u_i^k)^{\mathsf T} \mathbf y_i^k \ge b > 0$.

\emph{Step 3: Pairwise Contraction Map.}
For any non-zero vectors $\mathbf a, \mathbf c \in \mathbb R^2$, the unit projection distance obeys:
\begin{equation}\label{eq:norm_projection_ineq}
 \norm{\frac{\mathbf a}{\norm{\mathbf a}} - \frac{\mathbf c}{\norm{\mathbf c}}} \le \frac{\norm{\mathbf a - \mathbf c}}{\min(\norm{\mathbf a}, \norm{\mathbf c})}.
\end{equation}
Applying inequality~\eqref{eq:norm_projection_ineq} to updated headings $\mathbf u_i^{k+1} = \mathcal N_i(\mathbf y_i^k)$ and $\mathbf u_j^{k+1} = \mathcal N_j(\mathbf y_j^k)$ yields:
\begin{equation}\label{eq:proof_step3_contract1}
 \norm{\mathbf u_i^{k+1} - \mathbf u_j^{k+1}} \le \frac{\norm{\mathbf y_i^k - \mathbf y_j^k}}{\min(\norm{\mathbf y_i^k}, \norm{\mathbf y_j^k})} \le \frac{\norm{\mathbf y_i^k - \mathbf y_j^k}}{b}.
\end{equation}
Evaluating the numerator difference using decomposition~\eqref{eq:proof_step2_decomp}:
\begin{equation*}
 \mathbf y_i^k - \mathbf y_j^k = (1 - \lambda)(\mathbf u_i^k - \mathbf u_j^k) + (\mathbf E_i^k - \mathbf E_j^k).
\end{equation*}
Using triangle inequality and bound~\eqref{eq:proof_step2_Ebound}:
\begin{align*}
 \norm{\mathbf y_i^k - \mathbf y_j^k} &\le (1 - \lambda)\norm{\mathbf u_i^k - \mathbf u_j^k} + \norm{\mathbf E_i^k} + \norm{\mathbf E_j^k} \nonumber \\
 &\le (1 - \lambda)\Delta_k + \frac{4\lambda \Delta_k^3}{\epsilon_{\min}^2}.
\end{align*}
Substituting into~\eqref{eq:proof_step3_contract1} and maximizing over all pairs $i, j$:
\begin{equation*}
 \Delta_{k+1} \le \frac{(1 - \lambda) + \frac{4\lambda \Delta_k^2}{\epsilon_{\min}^2}}{b} \Delta_k.
\end{equation*}
Since $\Delta_k \le \Delta_0 \le \delta$, the coefficient is bounded by $q$ in~\eqref{eq:q_def}, yielding $\Delta_{k+1} \le q \Delta_k$. By induction, $\Delta_k \le q^k \Delta_0$ for all $k \ge 0$.

\emph{Step 4: Spatial Diameter Convergence.}
Using update rule~\eqref{eq:position}, position increments satisfy:
\begin{equation*}
 \norm{\mathbf p_i^{k+1} - \mathbf p_j^{k+1}} \le \norm{\mathbf p_i^k - \mathbf p_j^k} + v_0 \Delta t \norm{\mathbf u_i^{k+1} - \mathbf u_j^{k+1}}.
\end{equation*}
Maximizing over pairs gives $X_{k+1} \le X_k + v_0 \Delta t \, \Delta_{k+1} \le X_k + v_0 \Delta t \, \Delta_0 q^{k+1}$. Telescoping the summation from $0$ to $k-1$:
\begin{equation*}
 X_k \le X_0 + v_0 \Delta t \, \Delta_0 \sum_{m=1}^k q^m \le X_0 + v_0 \Delta t \, \Delta_0 \frac{q}{1 - q}.
\end{equation*}
Since $q < 1$, heading increments $\norm{\mathbf u_i^{k+1} - \mathbf u_i^k} \le 2\lambda \Delta_k$ are summable, guaranteeing heading convergence $\mathbf u_i^k \to \mathbf u_\star$. Relative position increments satisfy $\norm{(\mathbf p_i^{k+1}-\mathbf p_j^{k+1})-(\mathbf p_i^k-\mathbf p_j^k)}\le v_0\Delta t\,\Delta_{k+1}$ and are summable, so relative positions converge.
\end{proof}

\begin{remark}[Admissible Step-Size Region]
While Proposition~\ref{prop:safe} requires $\lambda < 1/2$ for arbitrary global states ($\Delta_k = 2$), local small-spread conditions ($\Delta_k \le \delta$) enlarge the admissible step-size range to $0 < \lambda \le 1$, provided the contraction condition $q < 1$ in~\eqref{eq:q_def} holds.
\end{remark}

\subsection{Offline Spatial Bootstrap Certificate}
Theorem~\ref{thm:local} assumes complete interaction and distinct positions along the trajectory. Proposition~\ref{prop:bootstrap} provides an offline-verifiable initial condition certifying these properties a priori.

\begin{proposition}[A Priori Bootstrap Condition]\label{prop:bootstrap}
Under the numerical conditions of Theorem~\ref{thm:local}, set cumulative relative spatial drift $$S = v_0 \Delta t \, \Delta_0 \frac{q}{1 - q}.$$ If initial positions satisfy:
\begin{equation*}
 X_0 + S < R, \qquad \min_{i \neq j} \norm{\mathbf p_i^0 - \mathbf p_j^0} > S,
\end{equation*}
then the graph remains complete ($g_{ij}^k = 1, \, \forall i \neq j$) and agent positions remain strictly distinct ($\norm{\mathbf p_i^k - \mathbf p_j^k} > 0$) for all $k \ge 0$, fully validating Theorem~\ref{thm:local}.
\end{proposition}
\begin{proof}
Suppose $k^* \ge 1$ is the earliest time step where either complete interaction fails or a pairwise collision occurs. For all $k < k^*$, Theorem~\ref{thm:local} holds, bounding cumulative pairwise spatial displacement by $\sum_{m=1}^{k^*} v_0 \Delta t \Delta_m \le S$. 

By the triangle inequality, every pairwise distance at $k^*$ satisfies:
\begin{equation*}
 \norm{\mathbf p_i^{k^*} - \mathbf p_j^{k^*}} \le \norm{\mathbf p_i^0 - \mathbf p_j^0} + S \le X_0 + S < R,
\end{equation*}
preventing edge loss, and:
\begin{equation}
 \norm{\mathbf p_i^{k^*} - \mathbf p_j^{k^*}} \ge \norm{\mathbf p_i^0 - \mathbf p_j^0} - S \ge \min_{l \neq m}\norm{\mathbf p_l^0 - \mathbf p_m^0} - S > 0,
\end{equation}
preventing coincidence. This contradicts the existence of $k^*$, proving completeness and distinctness for all $k \ge 0$.
\end{proof}

\subsection{Linearization Limits and Invariant Polarized Counterexample}
Linearizing heading dynamics near consensus ($\mathbf u_i = \mathbf u_\star + \mathbf t_\star z_i + O(\norm{\mathbf z}^2)$) yields scalar stochastic consensus $z_i^+ = (1 - \frac{\lambda d_i}{1+d_i})z_i + \frac{\lambda}{1+d_i}\sum_j g_{ij}z_j$, because gate departures $1 - h_\epsilon(s) = O(s^2)$ are higher order. First-order linearization cannot reveal component-selective gating.

Matrix graph connectivity alone is insufficient to guarantee global flocking consensus. To demonstrate this, consider $R = \infty$, four agents at positions $\mathbf p = (-3, -2, 2, 3) \times \{0\}$, and headings $\mathbf u_1 = \mathbf u_2 = (-1, 0)^{\mathsf T}$, $\mathbf u_3 = \mathbf u_4 = (1, 0)^{\mathsf T}$. With thresholds $\epsilon_r = 0.5, \epsilon_t = 0.3$, cross-camp relative headings are horizontal ($\mathbf w = (2, 0)^{\mathsf T}$), yielding $\mathbf P^r \mathbf w = (2, 0)^{\mathsf T}$ (rejected) and $\mathbf P^t \mathbf w = \mathbf 0$. Cross-camp matrix blocks equal $\mathbf B_{ij} = \operatorname{diag}(0, 1)$, while within-camp blocks equal $\mathbf I_2$. Cross-camp corrections are annihilated by $\mathbf B_{ij}$, and within-camp differences vanish. Every proposal satisfies $\mathbf y_i^k = \mathbf u_i^k$. The camps drift apart indefinitely while the matrix block graph remains connected (having a 3D kernel in its block Laplacian), demonstrating an exact invariant polarized equilibrium.

\subsection{Graph Connectivity and Clustered Consensus}

To analyze the structural properties of matrix-weighted networks, let $\mathcal{G} = (\mathcal{V}, \mathcal{E}, \mathbf{B})$ be an undirected matrix-weighted graph on $n$ agents with symmetric block edge weights $\mathbf{B}_{ij} = \mathbf{B}_{ji} \succeq \mathbf{0}$ in $\mathbb{R}^{d \times d}$ (where $d = 2$). The symmetric block Laplacian $\mathbf{L}_B \in \mathbb{R}^{nd \times nd}$ associated with $\mathcal{G}$ is defined block-wise as:
\begin{equation*}
 [\mathbf{L}_B]_{ij} = \begin{cases} 
 -g_{ij} \mathbf{B}_{ij}, & \text{if } i \neq j, \\
 \sum_{l \neq i} g_{il} \mathbf{B}_{il}, & \text{if } i = j.
 \end{cases}
\end{equation*}

\begin{definition}[Matrix Graph Connectivity~\cite{trinh2018}]
The matrix-weighted graph $\mathcal{G}$ is said to be \emph{matrix-algebraically connected} if the kernel (null space) of its block Laplacian $\mathbf{L}_B$ satisfies:
\begin{equation*}
 \operatorname{ker}(\mathbf{L}_B) = \operatorname{span}\left( \mathbf{1}_n \otimes \mathbf{I}_d \right) = \left\{ \mathbf{1}_n \otimes \mathbf{v} : \mathbf{v} \in \mathbb{R}^d \right\}. 
\end{equation*}
Equivalently, $\mathcal{G}$ is matrix-connected if the algebraic multiplicity of the zero eigenvalue of $\mathbf{L}_B$ is exactly $d$ (i.e., $\dim(\operatorname{ker}(\mathbf{L}_B)) = d$), and its $nd - d$ remaining eigenvalues are strictly positive ($\lambda_{d+1}(\mathbf{L}_B) > 0$).
\end{definition}

In standard linear matrix-weighted consensus $\dot{\mathbf{x}} = -\mathbf{L}_B \mathbf{x}$, matrix-algebraic connectivity ($\lambda_{d+1}(\mathbf{L}_B) > 0$) is both necessary and sufficient for all agents to converge to a global consensus vector $\mathbf{x}_1 = \dots = \mathbf{x}_n \in \mathbb{R}^d$~\cite{trinh2018}.

In the counterexample presented in Section~\ref{sec:analysis}-D, the underlying graph exhibits $\dim(\operatorname{ker}(\mathbf{L}_B)) = 3 > d = 2$. Specifically, the block Laplacian kernel decomposes into:
\begin{equation*}
 \operatorname{ker}(\mathbf{L}_B) = \operatorname{span}\left( \mathbf{1}_n \otimes \begin{bmatrix} 0 \\ 1 \end{bmatrix}, \, \mathbf{c}_1 \otimes \begin{bmatrix} 1 \\ 0 \end{bmatrix}, \, \mathbf{c}_2 \otimes \begin{bmatrix} 1 \\ 0 \end{bmatrix} \right),
\end{equation*}
where $\mathbf{c}_1 = (1, 1, 0, 0)^{\mathsf T}$ and $\mathbf{c}_2 = (0, 0, 1, 1)^{\mathsf T}$ are indicator vectors for the two spatial camps. 

This structure directly reflects the \emph{clustered consensus} characterization for PSD matrix weights:
\begin{enumerate}
 \item \emph{Orthogonal Channel Splitting:} The tangential channel ($\mathbf{P}^t = \operatorname{diag}(0, 1)$) remains globally connected across the network ($\lambda_2(\mathbf{L}_{B,y}) > 0$), holding a common vertical mode.
 \item \emph{Radial Channel Fragmentation:} The state-dependent radial gates yield $a^r_{ij} = 0$ for cross-camp edges, causing the radial matrix-weighted Laplacian $\mathbf{L}_{B,x}$ to disconnect into two uncoupled cluster components ($\mathbf{c}_1, \mathbf{c}_2$).
\end{enumerate}
Because the state-dependent gates $h_\epsilon$ create orthogonal rank-deficient projections ($\mathbf{B}_{ij} = \operatorname{diag}(0, 1)$), the nonlinear update map restricts vector differences to an invariant subspace manifold where cross-camp heading corrections vanish identically. Thus, unlike scalar-weighted networks where spatial connectivity implies scalar graph connectivity, state-dependent PSD matrix weights permit persistent \emph{clustered or polarized consensus} even when the scalar communication graph is complete.

\section{Numerical Results} \label{sec:numerical}
In this section, we present numerical simulations to validate the theoretical bounds in Section~\ref{sec:analysis} and evaluate performance against baselines. Unless stated otherwise, the principal benchmarks consider $n = 100$ agents with interaction radius $R = 8$, constant speed $v_0 = 0.1$, update interval $\Delta t = 1$, relaxation parameter $\lambda = 0.4$ (satisfying $0 < \lambda < 1/2$), gating thresholds $\epsilon_r = \epsilon_t = 0.85$, zero noise ($\eta = 0$), and 300 updates. 

Seed 7 initializes a two-camp perturbed ring: equally spaced bearings, radii uniform on $[2.6, 3.4]$, and Gaussian position perturbations with standard deviation $0.12$. Half of the initial headings are centered at $0.1$~rad and half at $1.4$~rad, each with a Gaussian perturbation of standard deviation $0.04$~rad. Population checks maintain this spatial footprint and interaction radius and evaluate increased density for $n \in \{10, 50, 100, 200\}$.

\begin{figure}[t]
\centering\includegraphics[width=0.8\columnwidth]{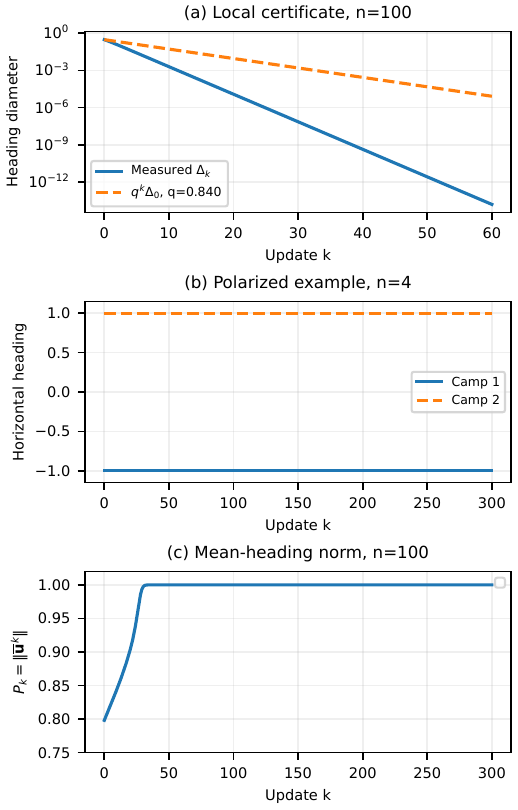}
\caption{Validation of theoretical diagnostics with independent initial conditions. (a)~Local heading contraction for $n=100$ agents ($\Delta_0=0.2988763, \delta=0.3$) versus the linear envelope $q^k \Delta_0$ predicted by Theorem~\ref{thm:local}. (b)~Exact four-agent invariant polarized trajectory ($R=\infty$). (c)~Evolution of the mean heading norm $P_k$ under state normalization ($n=100$).}
\label{fig:theory}
\end{figure}

\emph{Verification of Well-Posedness Bounds (Proposition~\ref{prop:safe}):}
Proposition~\ref{prop:safe} asserts that $0 < \lambda < 1/2$ guarantees non-vanishing proposal vectors ($\norm{\mathbf y_i^k} > 1 - 2\lambda = 0.2$) across arbitrary states. In the principal 100-agent, 300-step directional run, the minimum proposal vector norm observed across all agents and time steps was $\min_{i,k} \norm{\mathbf y_i^k} = 0.964784$. The corresponding evolution of the order parameter / mean heading norm $P_k$ under state normalization is shown in Fig.~\ref{fig:theory}(c). This confirms the analytical lower bound, verifying that state normalization $\mathcal N_i(\mathbf y_i^k)$ encounters no division-by-zero singularities during system evolution.

\emph{Verification of Local Flocking and Spatial Bootstrap (Theorem~\ref{thm:local} \& Proposition~\ref{prop:bootstrap}):}
To test the local contraction theorem (Theorem~\ref{thm:local}) and its offline bootstrap certificate (Proposition~\ref{prop:bootstrap}) at a larger scale, we place $n = 100$ agents on a centered $10 \times 10$ square grid with spatial spacing $0.5$ and initialize heading angles uniformly over $[-0.15, 0.15]$~rad. The initial heading spread in dimensionless chord distance is $\Delta_0 = 2\sin(0.15) = 0.2988763 < \delta = 0.3$.
Evaluating the theoretical constants in Theorem~\ref{thm:local} yields $b = 0.952104$ and contraction factor $q = 0.839518 < 1$. With an initial spatial diameter $X_0 = 6.363961$ and minimum pairwise separation $0.5$, the cumulative spatial drift bound from Proposition~\ref{prop:bootstrap} is $S = 0.156349$.
Checking the a priori conditions in Proposition~\ref{prop:bootstrap} gives:
\begin{align*}
 &X_0 + S = 6.363961 + 0.156349 = 6.520310 < R = 8, \\&
 \min_{i \neq j} \norm{\mathbf p_i^0 - \mathbf p_j^0} = 0.5 > S = 0.156349.
\end{align*}
Both conditions hold strictly, certifying offline that the communication graph remains complete and agent positions remain distinct for all $k \ge 0$. As shown in Fig.~\ref{fig:theory}(a), the empirical heading diameter $\Delta_k$ over 60 updates stays strictly below the theoretical contraction envelope $q^k \Delta_0$, confirming Theorem~\ref{thm:local} for an enlarged population and initial spread.

\emph{Demonstration of Polarized Invariant States:}
To empirically validate the structural limit established in Section III-D, we simulate the exact four-agent counterexample setup. As shown in Fig.~\ref{fig:theory}(b), cross-camp relative headings remain purely radial ($\mathbf w = (2, 0)^{\mathsf T}$), yielding matrix blocks $\mathbf B_{ij} = \operatorname{diag}(0, 1)$ that eliminate cross-camp corrections. The two camps drift apart indefinitely, confirming the analytical prediction.

\subsection{Benchmarking and Population Sweeps}
We evaluate the directional confidence law against four baselines using $n=100$ two-camp setup (seed 7):
\begin{enumerate}
    \item \emph{Unfiltered Vicsek:} Standard neighborhood averaging with $\mathbf B_{ij} = \mathbf I_2$.
    \item \emph{Smooth Tied Confidence:} Scalar gate $\mathbf B_{ij} = h_{0.85}(\norm{\mathbf w_{ij}})\mathbf I_2$.
    \item \emph{Matched Scalar:} Trace-matched gain $\mathbf B_{ij} = \frac{1}{2}(a^r_{ij} + a^t_{ij})\mathbf I_2$ sharing the exact residual update and degree denominator.
    \item \emph{Restricted-Angle Model~\cite{romensky2014}:} Streaming-first, self-inclusive averaging with no relaxation ($\lambda = 1$) and angular bound $\alpha_0 = 2\arcsin(0.85/2) = 0.87792$~rad matching the chord support of the $0.85$ tied gate.
\end{enumerate}

Performance across all strategies is evaluated at step $k = 300$ using three quantitative metrics:
\begin{enumerate}
    \item \emph{Heading RMS Disagreement ($E_k$):} The root-mean-square deviation of agent headings from the instantaneous spatial mean heading $\overline{\mathbf u}^k$:
    \begin{equation}
        E_k = \left( \frac{1}{n} \sum_{i=1}^n \norm{\mathbf u_i^k - \overline{\mathbf u}^k}^2 \right)^{1/2} = \sqrt{1 - P_k^2}. \label{eq:metric_E}
    \end{equation}
    Heading consensus is declared when $E_{300} \le 0.02$. 

    \item \emph{Order Parameter ($P_k$):} The norm of the average heading vector $P_k = \norm{\overline{\mathbf u}^k} \in [0, 1]$, where $P_k = 1$ indicates complete consensus.

    \item \emph{Final Cluster Count ($K$):} Descriptive count of distinct terminal heading camps formed by connecting headings whose chord distance is at most $0.05$.
\end{enumerate}

\begin{table}[t]
\caption{Benchmark ($n=100$, Seed 7, $k=300$ Updates)}
\label{tab:results}\centering
\begin{tabular}{lrrr}\hline
Interaction law & $E_{300}$ & $P_{300}$ & $K$\\\hline
Unfiltered Vicsek  & $<10^{-12}$ & 1.00000 & 1\\
Smooth Tied ($\epsilon = 0.85$) & 0.60173 & 0.79870 & 2\\
Directional Confidence ($\epsilon_r = \epsilon_t = 0.85$) & $<10^{-12}$ & 1.00000 & 1\\
Matched Scalar  & $<10^{-12}$ & 1.00000 & 1\\
Restricted Angle~\cite{romensky2014} & 0.60173 & 0.79870 & 2\\\hline
\end{tabular}
\end{table}

\begin{figure}[t]
\centering\includegraphics[width=0.8\columnwidth]{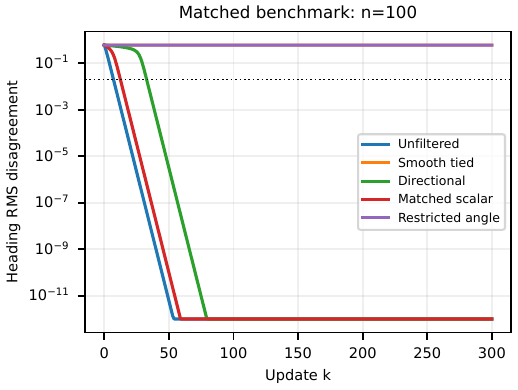}
\caption{Matched 100-agent benchmark trajectories and alignment residual evolution. The dotted horizontal line marks the alignment threshold ($0.02$); residuals below $10^{-12}$ are floored for display.}
\label{fig:comparison}
\end{figure}

As reported in Table~\ref{tab:results} and Fig.~\ref{fig:comparison}, directional confidence aligns the two-camp initial state ($E_{300} < 10^{-12}$), whereas smooth tied and restricted-angle confidence fail to align, remaining split into $K=2$ camps ($E_{300} = 0.60173$). Unfiltered and trace-matched scalar baselines also achieve complete alignment. 

Evaluating a diagnostic pair frame with $a^r=0$ and $a^t > 0$ demonstrates the component-selective matrix property: $\mathbf P^r\mathbf B_{ij}\mathbf w_{ij} = \mathbf 0$ while the accepted tangential correction remains active. A scalar gain rule cannot simultaneously suppress a rejected channel while preserving an orthogonal accepted channel on a vector with two non-zero components.

\begin{figure}[t]
\centering\includegraphics[width=0.8\columnwidth]{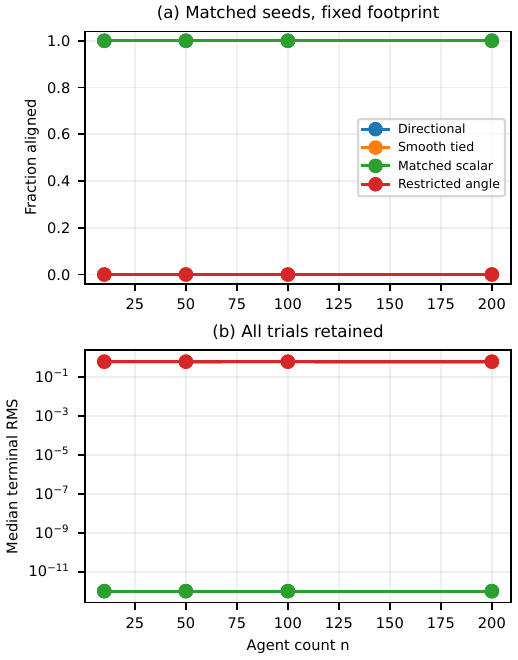}
\caption{Population density sweep over 20 matched seeds per size $n \in \{10, 50, 100, 200\}$: (a)~fraction aligned by update 300; (b)~median terminal RMS disagreement, floored at $10^{-12}$ for display. Increasing $n$ within a fixed footprint increases network density.}
\label{fig:size}
\end{figure}

To evaluate sensitivity to population density, we run matched-seed trials (seeds 0--19) across four network sizes $n \in \{10, 50, 100, 200\}$ under identical initial states. As shown in Table~\ref{tab:size} and Fig.~\ref{fig:size}, directional confidence and trace-matched scalar baselines achieve $100\%$ alignment success ($20/20$ trials) across all population sizes, whereas smooth tied and restricted-angle rules fail to align in all configurations.

\begin{table}[t]
\caption{Aligned Trials Out of 20 Matched Seeds per Population}
\label{tab:size}\centering
\begin{tabular}{rrrrr}\hline
$n$ & Directional & Tied & Matched & Restricted\\\hline
10 & 20 & 0 & 20 & 0\\
50 & 20 & 0 & 20 & 0\\
100 & 20 & 0 & 20 & 0\\
200 & 20 & 0 & 20 & 0\\\hline
\end{tabular}
\end{table}

\subsection{Empirical Robustness Under Near-Collision Initializations}
To evaluate numerical stability near the coincidence boundary ($\mathbf r_{12} = \mathbf 0$), we simulate two agents initialized at an extremely close spatial separation $\norm{\mathbf r_{12}^0} = 10^{-4}$ with initial headings $\mathbf u_1^0 = (1, 0)^{\mathsf T}$ and $\mathbf u_2^0 = (-0.6, 0.8)^{\mathsf T}$ ($\approx 127^\circ$ separation). 

Over 300 updates, the minimum observed proposal vector norm was $\min_{i,k} \norm{\mathbf y_i^k} = 0.998332$, staying well above the analytical lower bound $1 - 2\lambda = 0.20000$ (Proposition~\ref{prop:safe}). Unit speeds were preserved to floating-point precision ($\norm{\mathbf u_i^k} = 1$), confirming state normalization stability in close proximity.

\section{Discussion and Conclusion}\label{sec:conclusion}
In this paper, we introduced a directional, matrix-weighted bounded-confidence Vicsek model. By independently filtering radial and tangential relative heading components in the instantaneous LOS frame, the proposed residual alignment framework allows component-selective compatibility gating while guaranteeing that all synchronized agreement states remain invariant under zero noise.

Through a rigorous analysis of the discrete-time update map, we established global well-posedness by proving non-vanishing proposal vectors under appropriate step-size bounds, guaranteed linear geometric heading contraction for local small-spread configurations, and introduced an offline-verifiable spatial bootstrap condition that ensures continuous edge preservation and coincidence avoidance without relying on a priori topological assumptions.

The exact counterexample shows that connectivity of the graph of nonzero matrix blocks is insufficient for heading consensus; its block Laplacian has an extra null direction and is not matrix-algebraically connected. Numerical simulations confirmed all theoretical bounds, local contraction rates, and well-posedness margins.

Future theoretical work will focus on constructing non-local, state-dependent Lyapunov functionals to extend convergence analysis beyond small-spread configurations ($\Delta_k \le \delta$). A central mathematical challenge lies in characterizing anisotropic energy dissipation across rank-deficient matrix blocks $\mathbf{B}_{ij}$ while accounting for the invariant multi-camp equilibrium manifolds identified by our counterexample.

\section*{Acknowledgment}
OpenAI ChatGPT was used to assist with drafting, language refinement, and restructuring the
manuscript. The authors reviewed and revised the generated text and remain
responsible for all technical content.

\bibliographystyle{ieeetr}
\bibliography{references}

\vfill

\end{document}